\documentclass{article}

\PassOptionsToPackage{numbers, compress}{natbib}

\usepackage[main,final]{neurips_2026}

\usepackage[utf8]{inputenc} 
\usepackage[T1]{fontenc}    
\usepackage{float}          
\usepackage{placeins}       
\usepackage{hyperref}       
\usepackage{url}            
\usepackage{booktabs}       
\usepackage{amsfonts}       
\usepackage{nicefrac}       
\usepackage{microtype}      
\usepackage[table]{xcolor}  
\usepackage{multirow}       
\usepackage{anyfontsize}    
\usepackage{weibo_report}   

\definecolor{vibebg}{HTML}{FFF0F0}
\definecolor{vibered}{HTML}{D7192A}
\newcommand{\na}{--}
\newcommand{\our}[1]{\textbf{#1}}
\newcommand{\clrscore}[1]{\textbf{\textcolor{vibered}{#1}}}

\weibologo{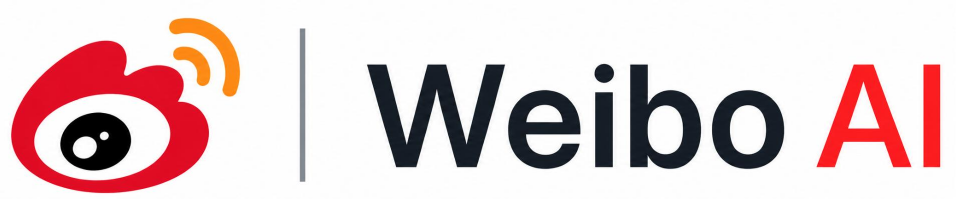}
\weiboheader{VibeThinker 3B Technical Report}
\title{VibeThinker-3B: Exploring the Frontier of Verifiable Reasoning in Small Language Models}

\weiboauthors{%
  Sen Xu,\enspace
  Shixi Liu,\enspace
  Wei Wang,\enspace
  Jixin Min,\enspace
  Yingwei Dai,\enspace
  Zhibin Yin,\enspace \\
  Yirong Chen,\enspace
  Xin Zhou,\enspace
  Junlin Zhang \enspace
}

\weiboaffils{%
  Sina Weibo Inc.
}

\weibodate{June 15, 2026}
\weibogithub{\texttt{https://github.com/WeiboAI/VibeThinker}}
\weibohuggingface{\texttt{https://huggingface.co/WeiboAI/VibeThinker-3B}}
\weibocorrespondence{\{xusen1,\,junlin6\} \texttt{@staff.weibo.com}}

\begin{document}

\maketitle
\enlargethispage{3\baselineskip}

\begin{abstract}
This technical report introduces VibeThinker-3B, a compact dense model with 3B parameters developed to investigate how far verifiable reasoning can be pushed within a strictly small-model regime. Building upon the Spectrum-to-Signal post-training paradigm, we systematically enhance the model through an optimized pipeline that includes curriculum-based supervised fine-tuning, multi-domain reinforcement learning, and offline self-distillation. Experimental evaluations demonstrate that VibeThinker-3B achieves frontier-level performance on highly demanding verifiable tasks. Specifically, it attains a score of 94.3 on AIME26 (improving to 97.1 with claim-level test-time scaling), an 80.2 Pass@1 on LiveCodeBench v6, and exhibits strong out-of-distribution generalization with a 96.1\% acceptance rate on recent unseen LeetCode contests. This effectively places it in the performance band of first-tier reasoning systems, matching or exceeding flagship models that are orders of magnitude larger, such as DeepSeek V3.2, GLM-5, and Gemini 3 Pro. Furthermore, a score of 93.4 on IFEval confirms that this extreme reasoning enhancement does not compromise strict instruction controllability. Extending our previous 1.5B work, these findings motivate the Parametric Compression-Coverage Hypothesis, which views verifiable reasoning as compressible into compact reasoning cores, while open-domain knowledge and general-purpose competence require broad parameter coverage over facts, concepts, and long-tail scenarios. This perspective suggests that compact models are not merely deployment-efficient substitutes, but a complementary path toward frontier-level performance in parameter-dense capability regimes.
\end{abstract}

\begin{figure}[H]
    \centering
    \includegraphics[width=0.98\textwidth]{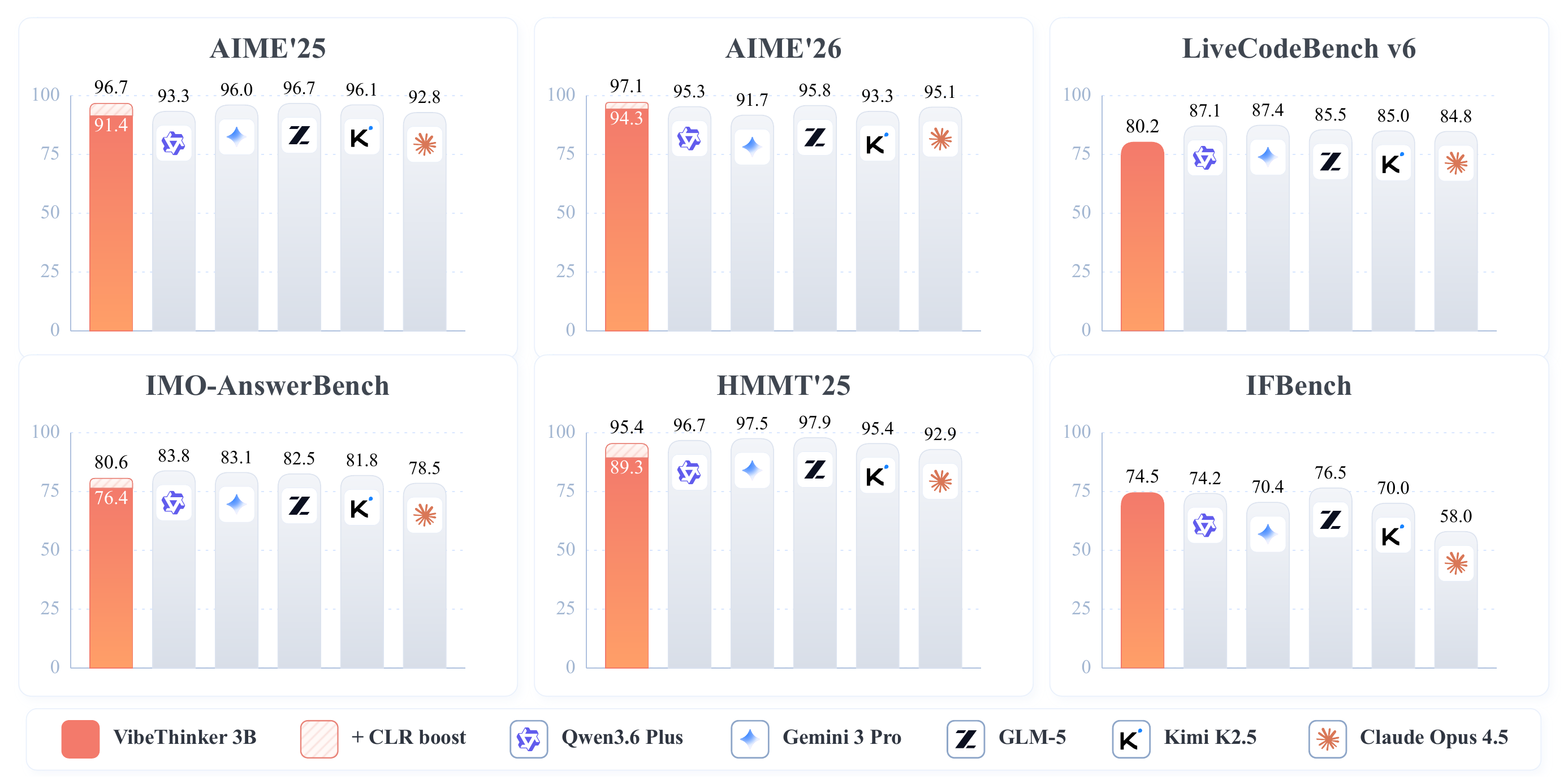}
    \vspace{-1ex}
    \caption{VibeThinker-3B reaches frontier reasoning performance at 3B scale. CLR denotes Claim-Level Reliability Assessment, a claim-level test-time scaling strategy.}
    \label{fig:main-results}
\end{figure}

\begin{figure}[H]
    \centering
    \includegraphics[width=0.92\textwidth]{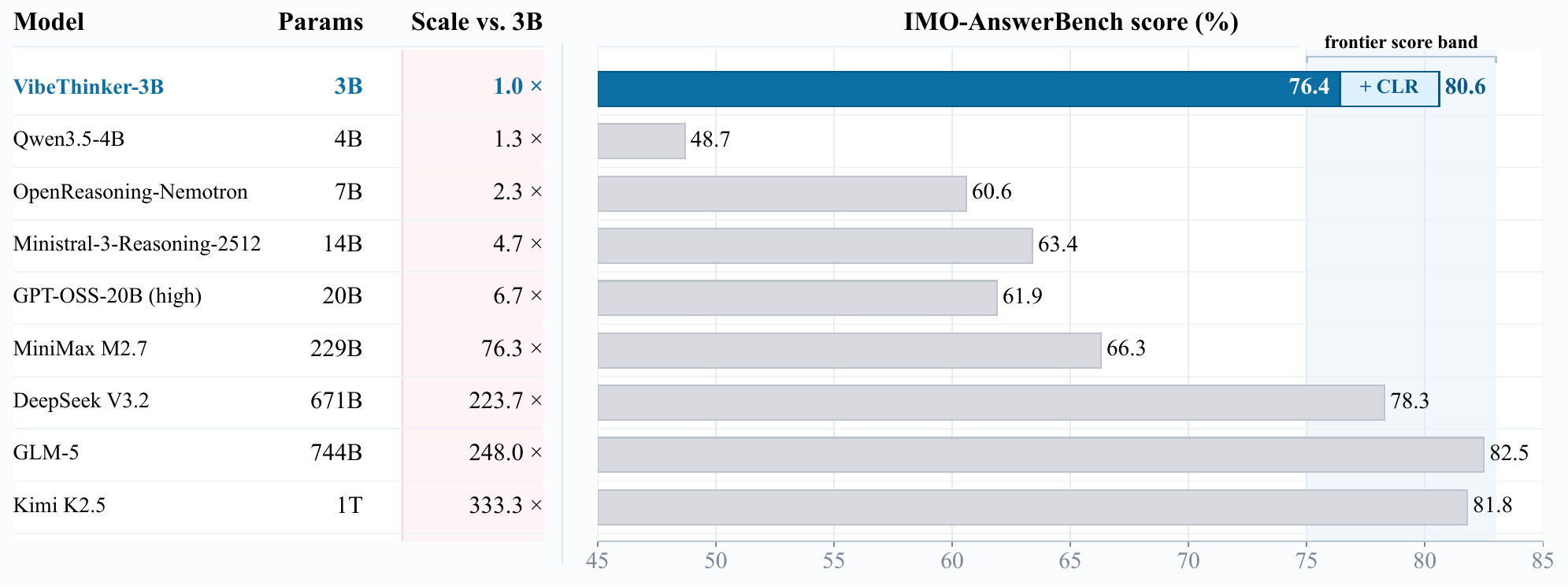}
    \vspace{-1ex}
    \caption{Parameter efficiency on IMO-AnswerBench, a highly demanding benchmark comprising 400 IMO-level problems, among open-source reasoning models with disclosed parameter counts. VibeThinker-3B achieves 76.4 with only 3B parameters and reaches 80.6 with CLR, demonstrating that a model within a strictly small-model regime can reach the performance range of substantially larger models such as DeepSeek V3.2 (78.3, 671B), GLM-5 (82.5, 744B), and Kimi K2.5 (81.8, 1T).}
    \label{fig:parameter-efficiency}
\end{figure}

\section{Introduction}

As reinforcement learning~\cite{shao2024GRPO,yu2025dapo,zheng2025gspo,hu2025reinforce++} has become increasingly integrated into the post-training stage of language models, the complex logical reasoning abilities of large models have improved substantially. At present, the field commonly relies on increasing parameter scale, following scaling laws, to cross the threshold required by difficult reasoning tasks. As a result, frontier reasoning ability is often concentrated in models with tens or hundreds of billions of parameters. In contrast, small language models (SLMs) with 3B parameters or fewer offer clear advantages in deployment cost, inference efficiency, and broader accessibility for academic research, but they are generally considered to face inherent bottlenecks when handling difficult mathematical derivations or complex programming tasks.

Our previous work on VibeThinker-1.5B~\cite{xu2025vibethinker} demonstrated that even models with extremely small parameter counts can be elicited to produce stable and basic chains of logic. This was an initial attempt to challenge the common belief that small models struggle with long-horizon reasoning. However, the 1.5B model mainly demonstrated the feasibility of reasoning in small models, while its upper bound remained to be explored. This led us to a further question: \textbf{\textit{instead of treating SLMs simply as compute-saving fallbacks, what is their true capability boundary? Can a strictly 3B model actually achieve frontier-level performance comparable to top-tier LLMs?}} Therefore, in this report, we present empirical observations on VibeThinker-3B to further examine the limits of complex verifiable reasoning at the 3B scale.

To further unlock the reasoning capacity of a 3B model, we systematically
upgrade the post-training pipeline built upon the Spectrum-to-Signal Principle introduced in VibeThinker-1.5B. In the SFT stage, we strengthen data synthesis,
quality filtering, and curriculum learning, allowing the model to first acquire
broad coverage across mathematics, code, STEM, general dialogue, and instruction
following, and then focus on harder long-horizon reasoning samples. In the RL
stage, we retain the core idea of MGPO~\cite{xu2025vibethinker} while extending training to multiple
verifiable domains and adopting a more stable long-context strategy to preserve
complete reasoning trajectories. We further introduce Long2Short Math RL to
improve reasoning efficiency by reducing redundant tokens without sacrificing
accuracy. Finally, offline self-distillation and Instruct RL consolidate the
capabilities elicited at different stages into a unified model and improve its
controllability under complex, constraint-heavy user instructions. Compared with
VibeThinker-1.5B, VibeThinker-3B therefore represents not only a moderate
increase in parameter scale, but also a more complete post-training system that
jointly addresses capability construction, reasoning amplification, efficiency
optimization, and instruction alignment.

Extensive evaluations across multiple independent competition systems, under strict data decontamination, confirm the exceptional parameter efficiency of VibeThinker-3B and ensure our findings are not isolated to a single benchmark. While it consistently outperforms existing small and mid-sized reasoning models, its most significant achievement is demonstrating competitiveness against top-tier systems that are orders of magnitude larger. Despite having only 3B parameters, VibeThinker-3B achieves a score of 94.3 on AIME26~\cite{maa2026aime}, matching the performance of much larger models such as DeepSeek V3.2~\cite{liu2025deepseekv3.2} (671B) and Kimi K2.5~\cite{team2026kimik2.5} (1T). It also scores 89.3 on HMMT25~\cite{hmmt2025} and achieves an 80.2 Pass@1 on LiveCodeBench~\cite{jain2024livecodebench} v6, closely trailing the performance of GPT-OSS-120B and DeepSeek V3.2. Furthermore, we employ Claim-Level Reliability Assessment (CLR), a test-time scaling strategy, which yields additional gains on answer-verifiable mathematics benchmarks, elevating its AIME26 score to 97.1, HMMT25 to 95.4, and BruMO25~\cite{brumo2025} to 99.2. Beyond standard benchmarks, VibeThinker-3B exhibits strong out-of-distribution generalization, achieving a 96.1\% acceptance rate on recent LeetCode weekly and biweekly contests (2026.04.25--2026.05.31), a level of pass rate comparable to industry-leading models such as GPT-5.2~\cite{openai2025gpt52systemcard} and Gemini 3 Flash~\cite{googledeepmind2025gemini3flashmodelcard}. 
Extending the technical lineage of the VibeThinker series, these achievements illustrate that a strict 3B parameter budget is entirely sufficient to approach the performance range of leading reasoning models such as Gemini 3 Pro, GLM-5, and Kimi K2.5, proving that the boundaries of reasoning capacity of compact models far exceeds conventional expectations.

Motivated by these findings, we introduce the Parametric Compression-Coverage Hypothesis, which posits that foundational model capabilities differ not only in the amount of parameter capacity they require, but also in the structural form of their parameter demands. Under this view, they can be broadly divided into parameter-dense capabilities and parameter-expansive capabilities. Verifiable reasoning exemplifies the former: its core challenge lies not in memorizing vast open-domain facts, but in performing search, constraint satisfaction, error correction, and multi-step composition within a structured solution space. Consequently, this class can be highly compressed into a compact and reusable reasoning core. In contrast, knowledge-intensive and general-purpose abilities align more closely with the latter, as they require broad coverage over open-domain facts, domain-specific concepts, semantic associations, and long-tail scenarios. Their parameter demands therefore resemble a coverage problem rather than the compression of a reusable reasoning core. This perspective elucidates why VibeThinker-3B achieves performance comparable to top-tier systems on verifiable tasks, such as mathematics and coding, while still exhibiting a gap relative to larger models on knowledge-intensive benchmarks such as GPQA-Diamond.

While parameter scaling remains a fundamental driver of broad model capabilities, we propose the Reasoning-Knowledge Decoupling Paradigm to reveal the highly specialized potential of smaller models. Under this paradigm, large-scale models continue to serve as natural vehicles for expansive knowledge breadth, as absorbing diverse semantics and long-tail distributions inherently requires massive parameter capacity. Conversely, provided with structurally constrained spaces and reliable training signals, smaller models are already sufficient to encapsulate high-density reasoning depth. Therefore, the true significance of VibeThinker-3B does not lie in proving that a 3B model can replace large-scale generalists, but rather in providing a concrete empirical signal: the development of compact models is no longer merely a passive compromise for deployment efficiency or cost control; it emerges as a promising research trajectory that is fundamentally complementary to the traditional parameter scaling paradigm.



\begin{figure}[t]
    \centering
    \includegraphics[width=\textwidth]{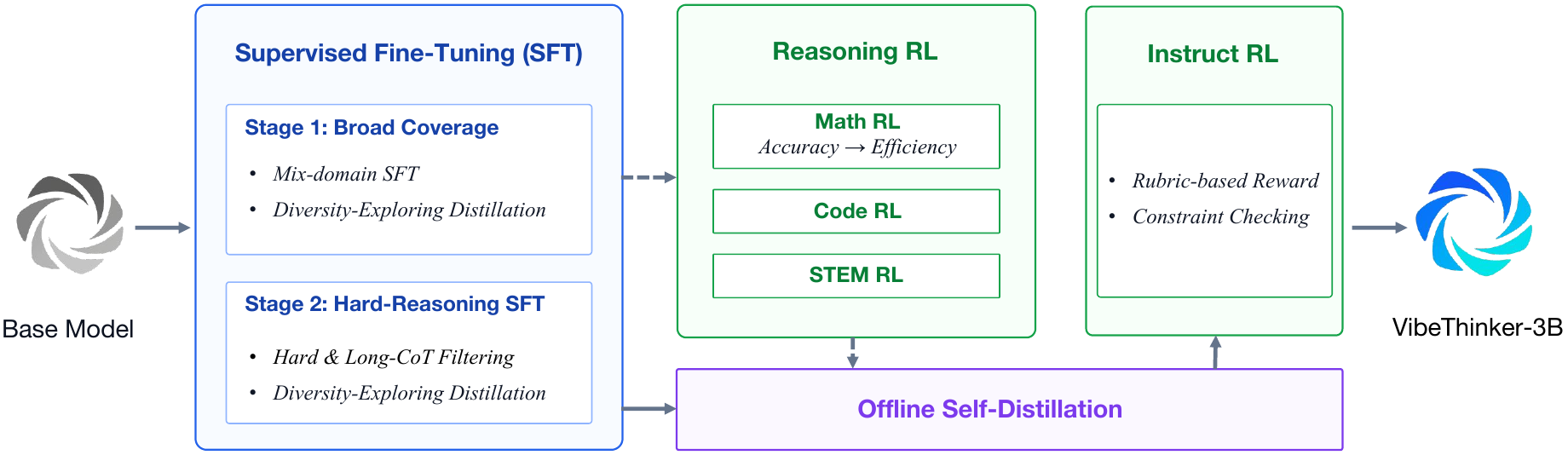}
    \caption{Overall training pipeline of VibeThinker-3B.}
    \label{fig1:pipeline}
\end{figure}

\section{Methods}

\textbf{Overall Pipeline.} VibeThinker-3B is developed through a staged post-training pipeline built upon Qwen2.5-Coder-3B base, a compact 3B dense foundation model. Our focus is on systematically eliciting and consolidating reasoning capabilities through data synthesis, diversity-oriented supervised fine-tuning, multi-domain reinforcement learning, offline self-distillation, and instruction-oriented alignment. The overall post-training framework continues the \textit{Spectrum-to-Signal Principle (SSP)} introduced in VibeThinker-1.5B ~\cite{xu2025vibethinker}. Building upon the core methodology of our previous work, we continue to employ Diversity-Exploring Distillation in the SFT stage to construct a broad solution space (the "Spectrum"), and utilize MaxEnt-Guided Policy Optimization (MGPO) in the RL stage to amplify high-value reasoning signals (the "Signal").

For this 3B iteration, we have comprehensively optimized the data construction and overall training pipeline based on our original foundation. As depicted in Fig.\ref{fig1:pipeline}, the complete post-training framework unfolds sequentially in stages. First, in the Supervised Fine-Tuning (SFT) stage, we have upgraded the rigorous data synthesis and filtering pipeline, thereby supporting the introduction of a two-stage curriculum learning strategy. This enables the model to transition smoothly from broad capability coverage to deep, long-horizon reasoning. Subsequently, in the Reinforcement Learning (RL) stage, we apply MGPO to multi-domain reasoning tasks utilizing a significantly expanded context window; furthermore, in the mathematical RL phase, we introduce a Long2Short stage designed to optimize reasoning efficiency without compromising accuracy. Following the completion of the core reasoning RL, the pipeline immediately proceeds to an Offline Self-Distillation phase to backfeed the newly elicited capabilities, and finally concludes with an Instruct RL stage to further reinforce the model's strict adherence to complex, multi-step instructions. The subsequent subsections will systematically elaborate on the detailed implementations of each stage.

\subsection{Supervised Fine-tuning}

\subsubsection{Data Construction}

During the SFT phase, we construct a multi-domain mixed supervised dataset based on the base model to provide a stable cold-start policy for subsequent RL phase. The dataset encompasses various tasks, including math, code, STEM reasoning, general chat, and instruction following. 

\textbf{Data Synthesis and Query Expansion.} VibeThinker-3B introduces an automated data synthesis pipeline during the SFT phase to broaden the coverage of training queries. We only select queries with reliable supervision signals from existing datasets as seed queries: mathematical queries must possess explicit and credible final answers or solving rationales, while competitive programming queries must be equipped with reliable unit tests or executable evaluation rules. Based on these high-confidence seed samples, we rewrite and expand the queries across multiple dimensions, e.g., concept composition, problem-solving skeletons, constraints, and evaluation objectives, yielding derivative queries that encompass a wider array of knowledge configurations and reasoning patterns. For the initially filtered synthetic queries, we further perform multiple independent samplings using strong teacher models and generate pseudo-labels via majority voting, establishing the foundation for subsequent distillation and training.

\textbf{Multi-path Reasoning Distillation.} For reasoning-intensive samples in mathematics, code, and STEM, we adopt a multi-path distillation approach to construct SFT responses. Specifically, we employ strong teacher models to sample multiple candidate reasoning traces for each query, retaining the complete intermediate reasoning steps rather than keeping only a single standard solution. This design inherits the Spectrum-to-Signal paradigm from VibeThinker-1.5B that the SFT phase is tasked with constructing a solution spectrum that covers diverse valid methods, offering a broader candidate solution space for subsequent RL. By explicitly preserving this multi-solution structure, the model learns various decomposition methods, derivation paths, and verification strategies, thereby improving exploration diversity during subsequent on-policy sampling.

\textbf{Multi-level Quality Control.} The quality of SFT data directly determines the performance upper bound of subsequent RL. Consequently, we implement more rigorous, multi-level quality control process:
\begin{itemize}
  \item \textit{1). N-gram-based filtering.} We discard samples containing anomalous repetitive segments, templated degeneration patterns, or n-gram overlaps with evaluation sets, to remove low-quality generations and benchmark contamination.
  \item \textit{2). LLM-based Query Quality Filtering.} We utilize capable LLMs to assess query quality, filtering out samples with incomplete descriptions, unreasonable conditions, invalid logic, or an inability to effectively assess target knowledge points.
  \item \textit{3). Trace Correctness Filtering.} At the distilled response level, we screen reasoning traces through a combination of answer verification, code sandbox execution, and LLM majority voting. Traces with incorrect final answers, failed execution results, or evidently invalid reasoning steps are filtered out.
\end{itemize}
The quality-controlled data is then stratified based on reasoning chain length and problem difficulty, establishing the data foundation for the subsequent curriculum SFT.

\subsubsection{Training Process}

\textbf{Curriculum-based two-stage SFT strategy.} Compared with VibeThinker-1.5B, VibeThinker-3B adopts a curriculum-based two-stage SFT procedure. The first stage focuses on broad capability coverage and behavioral cold start. We utilize the entire quality-filtered reasoning dataset for training to maximize the diversity of task types and reasoning patterns. Given the substantial variance in Chain-of-Thought (CoT) lengths within the first stage data, we employ sequence packing to enhance training efficiency. For optimization, we use a global batch size of 128 and set the initial learning rate to $5\times10^{-5}$. The learning rate follows a cosine annealing schedule and decays to a minimum value of $8\times10^{-8}$. The first stage is trained for 5 epochs with a 5\% linear warmup.

Upon acquiring a stable, broad-coverage SFT model, we proceed to the second stage, shifting the training data distribution toward higher-difficulty and longer-horizon reasoning samples. Initialized from the final checkpoint of the first stage, this phase continues training on a hard-reasoning subset generated through a joint length-difficulty filtration. Specifically, we first discard samples with reasoning traces shorter than 5K tokens. Subsequently, using VibeThinker-1.5B as a reference model, we perform 8 independent rollouts per query, filtering out relatively easy problems that yield an error rate below 0.75. This filtering strategy effectively reduces the proportion of shallow reasoning data, compelling the stage-two SFT to concentrate on long-horizon logical derivation, complex constraint satisfaction, and advanced problem-solving. Retaining the exact hyperparameter configuration from the first stage, this phase undergoes an additional 2 epochs of training on the hard-sample subset.

\textbf{Diversity-Exploring Distillation.} Following VibeThinker-1.5B~\cite{xu2025vibethinker}, we apply Diversity-Exploring Distillation in both SFT stages to mitigate potential gradient interference in multi-domain training and preserve the reasoning diversity of model outputs. This method follows the Spectrum-to-Signal Principle: the SFT stage does not aim for optimal imitation along a single solution path, but instead prioritizes the construction of a broader candidate solution space, providing a richer exploration basis for subsequent RL.

Specifically, we periodically save intermediate checkpoints during training and evaluate their Pass@K performance on domain-specific probing sets. For each domain, we select the checkpoint that produces more valid solutions as the corresponding specialist model, rather than simply choosing the checkpoint with the lowest validation loss or the highest Pass@1. These domain specialist models are then merged at the parameter level to obtain a unified SFT model. The resulting merged model preserves domain-specific reasoning capabilities while maintaining high output diversity, thereby providing a wider solution spectrum for subsequent training stages.

\subsection{Reinforcement Learning}
\subsubsection{Algorithm Backbone}

We reuse MaxEnt-Guided Policy Optimization (MGPO), introduced in VibeThinker-1.5B~\cite{xu2025vibethinker}, as the core RL algorithm. Under the Spectrum-to-Signal Principle, SFT constructs a diverse solution space, and RL is responsible for amplifying the correct reasoning signals within it. MGPO serves this role by dynamically selecting prompts near the model's current capability boundary.

For each prompt $q$, we sample $G$ responses from the old policy and evaluate them with verifiable rewards. The empirical group accuracy is computed as:
\begin{equation}
  p(q)=\frac{1}{G}\sum_{i=1}^{G}\mathbb{I}(r_i=1).
\end{equation}

Prompts with $p(q)\approx 0$ are too difficult and provide sparse positive signals, while prompts with $p(q)\approx 1$ are already saturated. Therefore, MGPO assigns higher weights to prompts with intermediate correctness:
\begin{equation}
  w(q)=\exp\left(-\gamma D_{\mathrm{ME}}(\,p(q)\,\|\,p_0\,)\right),\quad \mathrm{where}\,\, p_0 = 0.5,\, \gamma > 0.
\end{equation}

Here, $D_{\mathrm{ME}}(p(q)\|p_0)$ measures how far the empirical correctness $p(q)$ deviates from the maximum-entropy point $0.5$. A smaller value indicates that the prompt lies closer to the model's current capability boundary, where correct and incorrect rollouts coexist. This weight is applied to the group-relative advantage inside a GRPO-style clipped objective:
\begin{equation}
\mathcal{J}_{\mathrm{MGPO}}(\theta)
=
\mathbb{E}_{q,\{y_i\}}
\!\left[
\frac{1}{G}\sum_{i=1}^{G}
\frac{1}{|y_i|}\sum_{t=1}^{|y_i|}
\min\!\bigl(
\rho_{i,t}(\theta)\,w(q)A_i,\;
\mathrm{clip}(\rho_{i,t}(\theta),1\!-\!\varepsilon,1\!+\!\varepsilon)\,w(q)A_i
\bigr)
\right]\!,
\end{equation}

where $A_i$ is the group-relative advantage, $\rho_{i,t}(\theta)$ is the token-level probability ratio between the current and old policies, and $\varepsilon$ is the clipping coefficient. Inspired by the maximum-entropy principle, this weighting mechanism encourages RL updates to focus on prompts with sufficient uncertainty, thereby producing more stable and healthy gradient signals. It also mitigates over-optimization on high-probability tokens and reduces the negative impact of noisy tokens during policy updates.

In VibeThinker-3B, we keep the core MGPO formulation unchanged, while making several adjustments to improve the training stability. During training, we observe that as the rollout engine becomes increasingly optimized for inference throughput, the training-inference probability mismatch is gradually amplified by multiple implementation factors. This mismatch can destabilize or even collapse RL training. To mitigate this issue, we adopt the stabilization strategy from~~\cite{fangkhazi2025mismatch,yao2025mismatch} and perform all RL stages in an on-policy manner. 

\subsubsection{Multi-domain Reasoning RL}

We apply MGPO to multi-domain verifiable reasoning tasks, including mathematics, code, and STEM reasoning. These domains share the same policy optimization framework, but use different reward sources and verification mechanisms: mathematical tasks mainly rely on final-answer verification, code tasks rely on sandbox execution and test cases, and STEM tasks combine answer matching with option verification. 

\textbf{Training Data.} For all domains, the training sets comprise data with reliable supervision signals and have undergone strict benchmark decontamination. Additionally, before training commences, we filter out samples yielding an accuracy of exactly 0.0 or 1.0 as evaluated by the starting checkpoint of each respective phase.

\textbf{Single Long-context Learning.} ~\cite{luo2025deepscaler} introduce a multi-stage RL strategy based on progressive context-window expansion, improving both training efficiency and final reasoning performance. We observed a similar phenomenon in VibeThinker-1.5B, where progressively expanding the context window led to better reasoning performance with lower training cost. 

However, this conclusion does not hold in VibeThinker-3B. We find that a high-truncation early stage weakens the model's long-thinking capability and biases the policy toward incomplete or overly shortened reasoning trajectories. We hypothesize that this reversal is related to the stronger RL initialization checkpoint: compared with VibeThinker-1.5B, VibeThinker-3B undergoes stricter SFT data quality control and contains fewer invalid reasoning patterns. As a result, high-truncation warm-up may no longer mainly remove noisy thinking traces, but instead disrupt existing high-quality long-horizon reasoning behaviors. Even after the context window is expanded later, this degradation is difficult to fully recover. Therefore, we directly conduct RL with a single 64K long-context window, reducing rollout truncation and better preserving long-horizon reasoning trajectories.

\textbf{Training Strategy.} As illustrated in Fig.\ref{fig1:pipeline}, we adopt a sequential multi-domain Reasoning RL pipeline. Training starts with Math RL, which strengthens the model's long-horizon symbolic derivation, complex condition composition, and multi-step search capabilities. It then smoothly transitions to Code RL, focusing on improving the rigor of executable logic, boundary-case handling, and program constraint satisfaction. Finally, we conduct STEM RL to generalize the underlying logical reasoning ability to multidisciplinary scientific scenarios, enhancing knowledge utilization and cross-domain reasoning. The checkpoint obtained after each RL stage is preserved and used in the subsequent \textbf{\textit{offline self-distillation phase}}, where high-quality reasoning trajectories elicited at different stages are collected to further consolidate the model's overall reasoning capability.

\textbf{Long2Short Math RL.} Different from our previous work, VibeThinker-3B adopts a \textbf{\textit{'from accuracy to efficiency'}} two-stage reinforcement learning strategy. In the first stage, we optimize for accuracy using standard MGPO, allowing the model to fully unfold its reasoning process and explore diverse solution paths. Subsequently, we introduce a Long2Short stage in Math RL, extending the optimization objective from pure accuracy improvement to token-efficiency optimization. The goal of this stage is to reduce redundant reasoning and improve output efficiency while preserving validation-set performance. 
In Long2Short RL, we redistribute rewards only among correct trajectories in each prompt group according to response length, increasing the rewards of shorter correct responses and decreasing those of longer correct responses. After obtaining the binary correctness reward $r_i\in\{0,1\}$ for each sampled trajectory $y_i$, we keep all incorrect trajectories unchanged. For the correct set $\mathcal{C}=\{i\mid r_i=1\}$, we define a brevity score $s_i=1/L_i$, where $L_i$ denotes the response length, and apply a centered length-aware reward shift:
\[
r_i' = r_i + \lambda \cdot 
\frac{s_i-\bar{s}}{\max_{j\in\mathcal{C}} |s_j-\bar{s}|},
\qquad i\in\mathcal{C},
\]
where $\bar{s}$ is the mean brevity score over correct trajectories and $\lambda$, set to 0.2, controls the maximum redistribution magnitude. If all correct trajectories have the same length, the rewards are left unchanged. Since the reward shifts are centered within $\mathcal{C}$, their sum is zero:
\[
\sum_{i\in\mathcal{C}}(r_i'-r_i)=0.
\]
Therefore, the mean reward before and after redistribution remains unchanged for the correct subset and, since incorrect rewards are also unchanged, for the whole prompt group. This zero-sum design avoids introducing a systematic shift to the group-level reward baseline used in advantage estimation, while still reshaping the relative preference among correct trajectories toward more concise reasoning paths.

\subsection{Offline Self-Distillation}

After completing multi-domain Reasoning RL, we use the checkpoints from the Math, Code, and STEM RL stages, together with data filtering, to extract offline trajectories that contain high-quality reasoning patterns. These trajectories are then distilled back into a unified student model through supervised fine-tuning, enabling more stable integration of multi-domain reasoning capabilities.

\textbf{Learning-potential Filtering.} We first perform rejection sampling with domain-specific verifiers to remove incorrect trajectories. After obtaining verified teacher trajectories, we further introduce a learning potential score to estimate the distillation value of each correct trace for the student model. Specifically, for an input $q$ and a verified teacher trajectory $y$, we compute the length-normalized negative log-likelihood under the student model:
\begin{equation}
S_{\mathrm{LP}}(q,y)
=
-\frac{1}{|y|}
\sum_{t=1}^{|y|}
\log \pi_{\theta_{\mathrm{stu}}}(y_t \mid q,y_{<t}).
\end{equation}
A higher score indicates that the trace, although successfully generated and verified by the teacher, is not yet well modeled by the student, and therefore carries higher distillation value.

To prevent this score from being biased by sequence length or abnormal tokens, we do not rank traces globally. Instead, we compute priorities within domain-specific length buckets. Extremely short traces are excluded from score-based selection, as their average score can be dominated by a few abnormal tokens; extreme high-score outliers are also filtered to reduce the impact of format errors, distributional shifts, or noisy samples. Finally, we prioritize verified traces from the middle-to-high score range and mix the selected data across Math, Code, and STEM to construct the offline self-distillation dataset.

\subsection{Instruct RL}

We finally apply Instruct RL to convert the reasoning-enhanced checkpoint into a more reliable user-facing model. We train on a mixed instruction dataset containing format-sensitive prompts, long-context instructions, and general alignment examples. For samples with explicit constraints, rewards are computed by rule-based validators that check format, ordering, item count, keyword constraints, and task completion. For open-ended prompts, we use rubric-based reward models to evaluate helpfulness, coherence, instruction adherence, and redundancy. By combining constraint checking with rubric-based rewards under the same on-policy RL framework, Instruct RL reinforces strict controllability while preserving the reasoning ability obtained from previous stages.



\section{Evaluation}
\subsection{Evaluation Setup}

\textbf{Benchmarks.}
We evaluate VibeThinker-3B on a broad set of verifiable and instruction-oriented
benchmarks that cover mathematical reasoning, code generation, scientific
knowledge, and instruction following. For mathematics, we use AIME25~\cite{aops2025aime}, AIME26~\cite{maa2026aime},
HMMT25~\cite{hmmt2025}, BruMO25~\cite{brumo2025}, and IMO-AnswerBench~\cite{luong2025imo}
(abbreviated as IMO-Ans in tables), which together include recent competition-style
problems with different formats and difficulty profiles. For coding, we report
LiveCodeBench~\cite{jain2024livecodebench} v6 and OJBench~\cite{wang2025ojbench} as standard executable-code benchmarks. We further
include GPQA-Diamond~\cite{rein2023gpqa} to measure graduate-level scientific reasoning, and IFEval
and IFBench to evaluate whether the reasoning-enhanced model can still follow
explicit user constraints. In addition to these standard benchmarks, we evaluate
recent LeetCode weekly and biweekly contests as a practical out-of-distribution
test for algorithmic problem solving.

\textbf{Evaluation protocol.}
All VibeThinker-3B evaluations are performed with vLLM as the inference backend.
Unless otherwise specified, we use temperature 1.0, top-$p=0.95$, and
top-$k=-1$ for benchmark evaluation. We do not impose an additional output
length cap beyond the model's maximum generation length, allowing the model to
complete long reasoning trajectories when needed. For mathematical tasks, unlike the evaluation in VibeThinker-1.5B, we
jointly use math verify and LLM-as-judge to evaluate answer consistency. This is
particularly important for benchmarks such as IMO-AnswerBench, where final answers can
take more complex forms and rule-based symbolic verification alone may produce
unreliable judgments. For code tasks, correctness is determined by executing
the generated solution against the corresponding tests.

To ensure the stability of the evaluation metrics, we adopt different repeated sampling
strategies based on the problem scale of various benchmarks. Specifically, for mathematical
benchmarks, we report the mean Pass@1 over 64 independent generations, except for
IMO-AnswerBench where 16 independent generations are used. For knowledge and coding
benchmarks, we calculate the average performance using 16 and 8 independent generations,
respectively. Scores of comparison models are collected from their released reports,
public leaderboards, or official evaluation records when available.

\textbf{Test-time scaling with claim-level reliability assessment.}
We additionally evaluate VibeThinker-3B with Claim-Level Reliability Assessment (CLR), a test-time scaling strategy for answer-verifiable tasks. Unlike most test-time scaling methods that aggregate whole reasoning traces, CLR focuses on the important claims that affect key decisions during problem solving. It follows a streamlined two-stage procedure. First, using the exact same sampling parameters as our standard evaluation, the model generates $K=32$ candidate trajectories per problem and extracts $M=5$ decision-relevant claims alongside the final answer for each trajectory. Second, the model acts as its own self-verifier, attempting to falsify or validate these extracted claims to yield binary verdicts $v_{k,m} \in \{0, 1\}$. To heavily penalize trajectories containing flawed intermediate logic, CLR maps these verdicts into a nonlinear trajectory-level reliability score $r_k$:
\begin{equation}
    r_k = \left( \frac{1}{M} \sum_{m=1}^{M} v_{k,m} \right)^M
\end{equation}
Finally, candidate answers are clustered by equivalence, and the answer maximizing the reliability-weighted aggregation is selected:
\begin{equation}
    \mathrm{Score}(G) = \sum_{\{k \mid y_k \in G\}} r_k
\end{equation}
This claim-level assessment effectively reduces noise from long traces without updating model parameters. Compared to trace-level self-verification methods that require processing the entire verbose trajectory, CLR isolates critical logical anchors to significantly reduce token consumption while consistently improving Pass@1 performance. In our experiments, we independently execute this entire test-time scaling flow $8$ times and report the averaged Pass@1 performance as ``+ CLR'' in Table~\ref{tab:top-tier-reasoning-clr}.


\subsection{Evaluation Results}

\textbf{Overview.}
The central question of this evaluation follows directly from our previous
VibeThinker-1.5B study. That model demonstrated that small models can perform
reasoning tasks well, rather than merely producing shallow or unstable reasoning
traces. VibeThinker-3B takes the next step: instead of asking whether a small
model can reason at all, we ask how much parameter capacity is needed for a
small model to enter the performance band of first-tier reasoning systems.
After increasing the scale from 1.5B to 3B while preserving the
diversity-driven post-training paradigm, the results below suggest that the
reasoning capability of compact models is not linearly bounded by their
parameter scale, and a 3B model can move from "strong for its size" toward
genuine first-tier competitiveness.

\begin{table}[htbp]
\centering
\caption{Performance of VibeThinker-3B on Core Benchmarks}
\label{tab:vibethinker3b-core-benchmarks}

\fontsize{5.85pt}{7.45pt}\selectfont
\setlength{\tabcolsep}{1.75pt}
\renewcommand{\arraystretch}{1.23}
\resizebox{\linewidth}{!}{%
\begin{tabular}{@{}ll|ccccc|cc|c|cc@{}}
\toprule
\multicolumn{2}{c|}{\textbf{Model}} &
\multicolumn{5}{c|}{\textit{Mathematics}} &
\multicolumn{2}{c|}{\textit{Coding}} &
\multicolumn{1}{c|}{\textit{Knowledge}} &
\multicolumn{2}{c}{\textit{Instruction}} \\
\textbf{Name} & \textbf{Params} &
\textbf{AIME25} & \textbf{AIME26} & \textbf{HMMT25} & \textbf{BruMO25} & \textbf{IMO-Ans} &
\textbf{LCBv6} & \textbf{OJBench} &
\textbf{GPQA-D} &
\textbf{IFEval} & \textbf{IFBench} \\
\midrule
\addlinespace[1.5pt]
\multicolumn{12}{l}{\textit{Small Reasoning Models}} \\
\addlinespace[1pt]
SmolLM3 & 3B & 36.7 & 41.0 & 26.0 & 49.2 & 28.7 & 29.1 & 5.2 & 41.7 & 71.2 & 27.6 \\
Hunyuan-4B-Instruct & 4B & 66.5 & 57.7 & 35.2 & 62.7 & 39.6 & 46.8 & 12.1 & 61.1 & 76.6 & 26.5 \\
Qwen3-4B-Thinking-2507 & 4B & 81.3 & 79.0 & 55.5 & 77.7 & 51.6 & 55.2 & 17.9 & 65.8 & 87.4 & 52.9 \\
Qwen3.5-4B & 4B & 79.8 & 84.0 & 73.8 & 83.5 & 48.7 & 62.0 & 23.5 & 76.2 & 89.8 & 59.2 \\
Olmo-3-Think & 7B & 67.9 & 69.1 & 43.8 & 69.0 & 49.4 & 52.6 & 15.6 & 46.2 & 77.9 & 30.0 \\
Mimo7B-RL-0530 & 7B & 70.2 & 76.0 & 48.5 & 79.8 & 53.9 & 52.2 & 20.2 & 60.6 & 59.7 & 31.6 \\
OpenReasoning-Nemotron & 7B & 78.2 & 80.2 & 63.5 & 78.8 & 60.6 & 64.9 & 25.9 & 61.1 & 44.0 & 31.3 \\
Gemma-4-it & 12B & 72.9 & 77.5 & 63.3 & 80.4 & 54.9 & 72.0 & \na & 78.8 & 88.4 & 45.2 \\
Phi4-Reasoning-Plus & 14B & 68.4 & 73.6 & 50.3 & 66.5 & 46.2 & 56.8 & 14.4 & 81.9 & 84.9 & 51.7 \\
Ministral-3-Reasoning-2512 & 14B & 82.9 & 85.0 & 67.1 & 86.7 & 63.4 & 66.0 & 15.1 & 71.2 & 73.9 & 32.3 \\
\addlinespace[1.5pt]
\midrule
\addlinespace[1.5pt]
\multicolumn{12}{l}{\textit{Large Reasoning Models}} \\
\addlinespace[1pt]
GPT-OSS-20B (high) & 20B & 91.7 & 90.2 & 76.7 & 86.7 & 61.9 & 61.0 & \na & 71.5 & 92.8 & 65.0 \\
Nemotron-3-Nano & 30B & 89.1 & 90.1 & \na & \na & 70.4 & 68.3 & \na & 73.0 & 92.8 & 71.5 \\
GLM-4.5-Air & 106B & 83.3 & \na & 69.2 & 90.0 & \na & 70.7 & \na & 75.0 & 86.3 & 37.6 \\
Qwen3-235B-A22B-Thinking & 235B & 92.3 & \na & 83.9 & \na & 70.5 & 74.1 & 32.5 & 81.1 & 87.8 & 51.2 \\
LongCat Flash & 560B & 90.6 & \na & 83.7 & \na & \na & 79.4 & 40.7 & 81.5 & 86.9 & \na \\
GPT-5 Nano (high) & N/A & 85.2 & \na & 75.6 & 80.8 & \na & \na & \na & 71.2 & \na & \na \\
\addlinespace[1.5pt]
\midrule
\addlinespace[1.5pt]
\rowcolor{vibebg}
\textbf{VibeThinker-3B} & 3B &
\our{91.4} & \our{94.3} & \our{89.3} & \our{93.8} & \our{76.4} &
\our{80.2} & \our{38.6} & \our{70.2} & \our{93.4} & \our{74.5} \\
\bottomrule
\end{tabular}
}
\end{table}

\textbf{Core benchmark performance.}
Table~\ref{tab:vibethinker3b-core-benchmarks} summarizes the main evaluation
results across mathematics, coding, knowledge, and instruction following. The
upper block of the table compares VibeThinker-3B with small and mid-sized
reasoning models, including SmolLM3~\cite{bakouch2025smollm3}, Hunyuan-4B-Instruct~\cite{tencent2025hunyuan4binstruct}, Qwen3.5-4B~\cite{qwen3.5}, Olmo-3-Think~\cite{olmo2025olmo3},
Mimo7B-RL-0530~\cite{coreteam2025mimo7B}, OpenReasoning-Nemotron~\cite{majumdar2025openreasoningnemotron}, Gemma-4-12B-it~\cite{google2026gemma412b}, and Phi4-Reasoning-Plus~\cite{abdin2025phi4}.
On the mathematics suite, it reaches 91.4 on AIME25, 94.3 on AIME26, 89.3 on HMMT25,
93.8 on BruMO25, and 76.4 on IMO-AnswerBench. Compared to strong small reasoning
baselines ($<$14B), our model establishes a substantial performance lead.

Crucially, the coding and instruction-following results demonstrate that this enhancement is not confined to a specific family of mathematical benchmarks. VibeThinker-3B reaches 80.2 on LiveCodeBench v6 and 38.6 on OJBench, surpassing all models in Table~\ref{tab:vibethinker3b-core-benchmarks} on LiveCodeBench v6. Furthermore, it achieves 93.4 on IFEval and 74.5 on IFBench, confirming that the reasoning optimization process does not compromise instruction controllability. This is particularly significant, as a practical small reasoning model must not only solve competitive problems but also maintain reliable user-facing alignment after undergoing long-context reasoning RL and self-distillation.

The lower block of Table~\ref{tab:vibethinker3b-core-benchmarks} extends the comparison to
substantially larger reasoning models, namely GPT-OSS-20B (high)~\cite{agarwal2025gptoss},
Nemotron-3-Nano~\cite{blakeman2025nemotron3nano}, GLM-4.5-Air~\cite{zeng2025glm4.5glm4.5air}, Qwen3-235B-A22B-Thinking~\cite{qwen3technicalreport}, and LongCat Flash~\cite{team2025longcat}.
This serves as a more rigorous test for the central claim articulated in the Introduction: if reasoning ability on
verifiable tasks depends primarily on abstract search, constraint satisfaction,
and error correction rather than parametric memorization, then a carefully
optimized 3B model should be capable of challenging much larger systems on these tasks. Our empirical results substantiate this hypothesis. VibeThinker-3B achieves leading performance across multiple benchmarks when compared to reasoning models several times its size, outperforming or matching several 30B--560B open models. This demonstrates
that the 3B parameter budget is already sufficient to support highly compressed, long-horizon mathematical reasoning, provided the model is trained with appropriate exploration and verification signals.

At the same time, the table reveals a useful boundary. The gap to the strongest
large models is more visible on broad knowledge-heavy evaluation, especially
GPQA-Diamond, than on competition mathematics or executable coding. This echoes
the observation from VibeThinker-1.5B: compact models can acquire strong
reasoning procedures, but knowledge-intensive benchmarks still expose a clear
gap to large-parameter general models. This pattern is consistent with our
hypothesis that reasoning and knowledge storage are only partially coupled.
Compact models may still face capacity limits when broad domain knowledge must
be recalled directly, but they can nevertheless host a highly effective
reasoning engine for tasks with verifiable goals and structured solution spaces.

\begin{table}[htbp]
\centering
\caption{Performance of VibeThinker-3B on Core Benchmarks (Top-Tier Reasoning Models)}
\label{tab:top-tier-reasoning-clr}

\fontsize{6.2pt}{7.6pt}\selectfont
\setlength{\tabcolsep}{1.8pt}
\renewcommand{\arraystretch}{1.20}
\resizebox{\linewidth}{!}{%
\begin{tabular}{@{}ll|ccccc|cc|c|cc@{}}
\toprule
\multicolumn{2}{c|}{\textbf{Model}} &
\multicolumn{5}{c|}{\textit{Mathematics}} &
\multicolumn{2}{c|}{\textit{Coding}} &
\multicolumn{1}{c|}{\textit{Knowledge}} &
\multicolumn{2}{c}{\textit{Instruction}} \\
\textbf{Name} & \textbf{Params} &
\textbf{AIME25} & \textbf{AIME26} & \textbf{HMMT25} & \textbf{BruMO25} & \textbf{IMO-Ans} &
\textbf{LCBv6} & \textbf{OJBench} &
\textbf{GPQA-D} &
\textbf{IFEval} & \textbf{IFBench} \\
\midrule
\addlinespace[1.5pt]
\multicolumn{12}{l}{\textit{Open-Source Models}} \\
\addlinespace[1pt]
GPT-OSS (high) & 120B & 92.5 & 93.2 & 90.0 & 92.5 & 75.6 & 81.9 & 41.5 & 80.1 & 89.5 & 69.5 \\
MiMo v2 Flash & 309B & 94.1 & \na & 84.4 & \na & \na & 80.6 & \na & 83.7 & \na & 64.2 \\
MiniMax M2.7 & 229B & \na & 89.8 & \na & \na & 66.3 & \na & \na & 87.0 & \na & 76.0 \\
DeepSeek R1 0528 & 671B & 87.5 & \na & 79.4 & 92.5 & 60.8 & 68.7 & 33.6 & 81.0 & 79.1 & 39.6 \\
Qwen3.5-397B-A17B & 397B & \na & 91.3 & 94.8 & \na & 80.9 & 83.6 & \na & 88.4 & 92.6 & 76.5 \\
DeepSeek V3.2 & 671B & 93.1 & 94.2 & 90.2 & 96.7 & 78.3 & 80.8 & 48.4 & 82.4 & 92.6 & 60.7 \\
Kimi K2.5 & 1T & 96.1 & 93.3 & 95.4 & 98.3 & 81.8 & 85.0 & 54.7 & 87.6 & 93.9 & 70.0 \\
GLM-5 & 744B & 96.7 & 95.8 & 97.9 & \na & 82.5 & 85.5 & 55.0 & 86.0 & 92.6 & 76.5 \\
\addlinespace[1.5pt]
\midrule
\addlinespace[1.5pt]
\multicolumn{12}{l}{\textit{Proprietary Models}} \\
\addlinespace[1pt]
Gemini 2.5 Flash & N/A & 72.0 & \na & 64.2 & 83.3 & \na & 61.2 & 23.5 & 82.8 & 89.8 & 36.1 \\
OpenAI o3 (high) & N/A & 88.9 & \na & 77.5 & 95.8 & 61.1 & 75.8 & 25.4 & 83.3 & 92.1 & 69.3 \\
Gemini 2.5 Pro & N/A & 86.7 & \na & 82.5 & 90.0 & 68.2 & 72.5 & 38.9 & 86.4 & 90.8 & 48.7 \\
Grok 4 & N/A & 91.7 & \na & 90.0 & 95.0 & 73.1 & \na & \na & 87.5 & 88.0 & 53.7 \\
Claude Opus 4.5 & N/A & 92.8 & 95.1 & 92.9 & \na & 78.5 & 84.8 & \na & 87.0 & \na & 58.0 \\
GPT-5 (high) & N/A & 94.6 & \na & 88.3 & 91.7 & 76.0 & 84.5 & \na & 85.7 & \na & 73.1 \\
Qwen3.6 Plus & N/A & 93.3 & 95.3 & 96.7 & \na & 83.8 & 87.1 & \na & 90.4 & 94.3 & 74.2 \\
Gemini 3 Pro & N/A & 96.0 & 91.7 & 97.5 & 98.3 & 83.1 & 87.4 & 58.8 & 91.9 & \na & 70.4 \\
\addlinespace[1.5pt]
\midrule
\addlinespace[1.5pt]
\rowcolor{vibebg}
\textbf{VibeThinker-3B} & &
\our{91.4} & \our{94.3} & \our{89.3} & \our{93.8} & \our{76.4} &
\our{80.2} & \our{38.6} & \our{70.2} & \our{93.4} & \our{74.5} \\
\rowcolor{vibebg}
\quad + CLR & \multirow{-2}{*}{3B} &
\clrscore{96.7} & \clrscore{97.1} & \clrscore{95.4} & \clrscore{99.2} & \clrscore{80.6} &
 & & \clrscore{72.9} & & \\
\bottomrule
\end{tabular}
}
\end{table}

\textbf{Comparison with top-tier reasoning models.}
Table~\ref{tab:top-tier-reasoning-clr} then raises the comparison bar from general baselines to top-tier reasoning models, including open-source models such as GPT-OSS-120B (high)~\cite{agarwal2025gptoss}, MiMo v2 Flash~\cite{xiao2026mimov2flash}, MiniMax M2.7~\cite{chen2026minimaxM2.7}, DeepSeek R1 0528~\cite{deepseekai2025deepseekr1R10528}, Qwen3.5-397B-A17B~\cite{qwen3.5}, DeepSeek V3.2~\cite{liu2025deepseekv3.2}, Kimi K2.5~\cite{team2026kimik2.5}, GLM-5~\cite{zeng2026glm5}, and proprietary models such as Gemini 2.5 Flash~\cite{comanici2025gemini2.5proflash}, OpenAI o3~\cite{openai2025o3o4mini}, Gemini 2.5 Pro~\cite{comanici2025gemini2.5proflash}, Grok 4~\cite{xai2025grok4}, Claude Opus 4.5~\cite{anthropic2025claudeopus45systemcard}, GPT-5 (high)~\cite{singh2025gpt5gpt5nano}, Qwen3.6 Plus~\cite{qwen36plus}, and Gemini 3 Pro~\cite{googledeepmind2026gemini3pro}, making the comparison even more demanding. These systems represent the current flagship regime in reasoning capability and are backed by substantially larger model and training budgets.

VibeThinker-3B still performs competitively in this setting. Without CLR reasoning enhancement, its AIME26 score of 94.3 is comparable to DeepSeek V3.2 and Kimi K2.5, while its 93.8 on BruMO25 exceeds several much larger parameter models. We also report the effect of CLR on answer-verifiable mathematics benchmarks and GPQA-Diamond. With CLR, VibeThinker-3B improves to 96.7 on AIME25, 97.1 on AIME26, 95.4 on HMMT25, 99.2 on BruMO25, 80.6 on IMO-AnswerBench, and 72.9 on GPQA-Diamond. After using CLR, the model enters the top cluster of Table~\ref{tab:top-tier-reasoning-clr} on competition-style mathematics: it matches or exceeds many flagship open-source and proprietary systems on AIME25, AIME26, HMMT25, BruMO25, and IMO-AnswerBench.

This result does not imply that a 3B model has matched leading general-purpose systems in comprehensive capabilities (such as broad encyclopedic knowledge or open-domain instruction following). Rather, it provides an important and concrete proof: on well-constrained, verifiable reasoning tasks, first-tier performance is no longer the exclusive domain of ultra-large models, and a compact model of merely 3B parameters can equally earn its place. In this sense, VibeThinker-3B acts as a "parameter-scale probe" built upon the conclusion of VibeThinker-1.5B: if the 1.5B version proved that a small model could produce complete and logically coherent reasoning trajectories, then the 3B version further answers the critical question of "what parameter threshold is actually required to enter the top reasoning tier". This is precisely the core empirical signal that VibeThinker-3B aims to deliver---under appropriate post-training optimization, extreme reasoning capability is not strictly bounded by raw parameter scale.

The GPQA-Diamond result should be interpreted more conservatively. CLR raises
VibeThinker-3B from 70.2 to 72.9, but the model still trails the strongest
large-parameter systems by a visible margin on this knowledge-heavy benchmark.
This is consistent with our claim rather than a contradiction to it: the main
finding is not that a 3B model has fully replaced leading general-purpose
models, but that a small model can reach first-tier performance on many
verifiable reasoning tasks. These results suggest that, within such domains,
the decisive bottleneck is not always raw parameter count; high-quality
post-training, diverse solution exploration, reliable verification, and
effective test-time reasoning can jointly push a compact model into a much
higher capability regime.

\begin{table}[htbp]
\centering
\caption{OOD Generalization Test: LeetCode Weekly \& Biweekly Contests (Apr 25--May 31, 2026)}
\label{tab:vibethinker-3b-leetcode}

\fontsize{6.2pt}{7.6pt}\selectfont
\setlength{\tabcolsep}{2.2pt}
\renewcommand{\arraystretch}{1.20}
\resizebox{0.92\linewidth}{!}{%
\begin{tabular}{@{}l|cccccccc|c@{}}
\toprule
\multicolumn{1}{c|}{\textbf{Model}} &
\multicolumn{8}{c|}{\textit{LeetCode Contests (Python)}} &
\multicolumn{1}{c}{\textit{Aggregate}} \\
\textbf{Name} &
\textbf{BW181} & \textbf{BW182} & \textbf{BW183} &
\textbf{W499} & \textbf{W500} & \textbf{W502} & \textbf{W503} & \textbf{W504} &
\textbf{Overall} \\
\midrule
\addlinespace[1pt]
GPT-5.3-Codex & 16/16 & 16/16 & 16/16 & 16/16 & 16/16 & 16/16 & 16/16 & 16/16 & 100.0\%\, (128/128) \\
Gemini 3.1 Pro & 16/16 & 16/16 & 16/16 & 16/16 & 16/16 & 16/16 & 15/16 & 16/16 & 99.2\%\, (127/128) \\
Gemini 3 Flash & 16/16 & 16/16 & 12/16 & 16/16 & 16/16 & 16/16 & 16/16 & 16/16 & 96.9\%\, (124/128) \\
\rowcolor{vibebg}
\textbf{VibeThinker-3B} &
\our{16/16} & \our{16/16} & \our{12/16} & \our{16/16} &
\our{16/16} & \our{16/16} & \our{15/16} & \our{16/16} &
\our{96.1\%\, (123/128)} \\
GPT-5.2 & 15/16 & 16/16 & 15/16 & 16/16 & 15/16 & 14/16 & 16/16 & 15/16 & 95.3\%\, (122/128) \\
Doubao Seed 2.0 Pro & 16/16 & 16/16 & 14/16 & 16/16 & 14/16 & 15/16 & 14/16 & 16/16 & 94.5\%\, (121/128) \\
Qwen3-Max & 16/16 & 16/16 & 12/16 & 16/16 & 10/16 & 16/16 & 15/16 & 16/16 & 91.4\%\, (117/128) \\
Kimi K2.5 & 16/16 & 16/16 & 12/16 & 15/16 & 12/16 & 16/16 & 15/16 & 14/16 & 90.6\%\, (116/128) \\
Qwen3.5-397B-A17B & 16/16 & 16/16 & 12/16 & 13/16 & 15/16 & 15/16 & 15/16 & 13/16 & 89.8\%\, (115/128) \\
GPT-5 mini & 16/16 & 16/16 & 12/16 & 14/16 & 11/16 & 16/16 & 15/16 & 13/16 & 88.3\%\, (113/128) \\
Claude Opus 4.6 & 15/16 & 16/16 & 12/16 & 16/16 & 12/16 & 16/16 & 15/16 & 9/16 & 86.7\%\, (111/128) \\
Claude Sonnet 4.6 & 16/16 & 15/16 & 11/16 & 16/16 & 10/16 & 14/16 & 11/16 & 9/16 & 79.7\%\, (102/128) \\
Grok 4.1 Fast & 15/16 & 15/16 & 11/16 & 13/16 & 13/16 & 12/16 & 9/16 & 11/16 & 77.3\%\, (99/128) \\
GLM-5 & 15/16 & 14/16 & 12/16 & 14/16 & 8/16 & 16/16 & 12/16 & 7/16 & 76.6\%\, (98/128) \\
\bottomrule
\end{tabular}
}
\end{table}

\textbf{Generalization test on recent LeetCode contests.}
To further test coding generalization beyond curated benchmark suites, we
evaluate VibeThinker-3B on recent LeetCode weekly and biweekly contests from
Apr. 25 to May 31, 2026. As shown in Table~\ref{tab:vibethinker-3b-leetcode}, each model is evaluated using
Python-only one-shot generation. Each contest column contains four problems, and each problem is sampled with
four independent rollouts, yielding 16 first-attempt submissions per contest. A cell of $x/16$ therefore means
that $x$ of the 16 independent Python submissions passed all hidden tests on their first submission. Weekly
contests are denoted by ``W'' and biweekly contests by ``BW''; for example, W504 refers to LeetCode Weekly
Contest 504. We omit W501 because it does not have public LeetCode LLM ranking data, and include W504 as the latest available weekly contest in the public leaderboard used here.

Overall, VibeThinker-3B passes 123 out of 128 first-attempt Python submissions, corresponding to an overall acceptance rate of 96.1\%. This result is higher than GPT-5.2~\cite{openai2025gpt52systemcard}, Doubao Seed 2.0 Pro~\cite{bytedanceseed2026seed20release}, Qwen3-Max~\cite{qwen3max}, Kimi K2.5~\cite{team2026kimik2.5}, Qwen3.5-397B-A17B~\cite{qwen3.5}, and the Claude 4.6~\cite{anthropic2026claudeopus46systemcard} models under the same contest aggregation. It is also close to Gemini 3 Flash~\cite{googledeepmind2025gemini3flashmodelcard} and remains below only the strongest entries in the table. The contest results are useful because the tasks are recent, diverse, and execution-verified. They therefore provide a complementary view to
LiveCodeBench and OJBench: VibeThinker-3B is not merely fitting a static coding
benchmark distribution, but can solve fresh algorithmic problems under a
realistic competitive-programming evaluation protocol. This confirms the model's
robust out-of-distribution (OOD) generalization capabilities on unseen algorithms.

\section{Conclusion}

In this report, we present VibeThinker-3B, a compact reasoning model comprising only 3 billion parameters. On challenging verifiable reasoning benchmarks, including AIME26, HMMT25, IMO-AnswerBench, and LiveCodeBench v6, it delivers strong results and further demonstrates robust generalization on out-of-distribution LeetCode evaluations. Taken together, these evaluations show that VibeThinker-3B reaches a performance band comparable to representative frontier LLMs, such as GLM-5, Kimi K2.5, Gemini 3 Pro, and Claude Opus 4.5, providing evidence that small language models can effectively approximate frontier reasoning capabilities on highly complex verifiable tasks despite much smaller parameter scales.

Based on these findings, we propose the Parametric Compression-Coverage Hypothesis, suggesting a structural divergence in how foundational capabilities are encoded within the parameter space. Specifically, verifiable reasoning aligns more closely with parameter-dense core compression, whereas open-domain knowledge and general-purpose capabilities rely more heavily on the broad coverage afforded by model scale. Consequently, the potential for small models to achieve top-tier performance within specific capability domains has long been underestimated. Therefore, the development of SLMs should no longer be viewed merely as a passive choice driven by deployment efficiency. Instead, it serves as an efficient and complementary evolutionary trajectory alongside traditional scaling laws, offering novel insights for the design of future reasoning systems.

{\small
\bibliographystyle{unsrtnat}
\bibliography{references}
}







\end{document}